# Word Embeddings Inherently Recover the Conceptual Organization of the Human Mind

Victor Swift

University of Toronto

**Correspondence**: victor.swift@mail.utoronto.ca

**Acknowledgments:** I am grateful to S. De Deyne, T. Joyce, D. Nelson, and their colleagues for compiling and sharing the Word Association data used herein, and T. Mikolov et al. for distributing high quality pre-trained Word Embeddings.

Preprint. Under review.

**Machine learning is a means to uncover deep patterns from rich sources of data. Here, we find that machine learning can recover the conceptual organization of the human mind when applied to the natural language use of millions of people. Utilizing text from billions of webpages, we recover most of the concepts contained in English, Dutch, and Japanese, as represented in large scale Word Association networks. Our results justify machine learning as a means to probe the human mind, at a depth and scale that has been unattainable using self-report and observational methods. Beyond direct psychological applications, our methods may prove useful for projects concerned with defining, assessing, relating, or uncovering concepts in any scientific field.**

Word Embedding (WE) represents a class of machine learning techniques used to uncover the semantic structure of text corpora. State-of-the-art WE algorithms utilize neural networks to calculate the semantic relatedness of all words within a corpus on the basis of contextual interchangeability[1]. Thus, words that occur in the same contexts (i.e., co-occur with the same words) are deemed more similar than words that occur in different contexts. In this way, WE can represent the relative meaning of all the words within a language.

In recent years, WE has proven to reliably represent the meaning of individual words in various lines of study. Most notably, the position of WE word vectors has been shown to predict human responses on semantic tasks[2-7] and performance on cognitive tasks[8-11]. Moreover, the changing position of WE vectors over time has been found to correspond with historical changes in word meaning[12-14]. While these findings provide definitive support for WE as a means to model language, additional work is needed to establish whether WE can model the conceptual organization of the human mind.

The organization of consciously accessible information in the human mind has been traditionally investigated using Word Association (WA) studies[15]. In such studies, individuals are presented with a cue word (e.g., *salty*) and asked to respond with the first word that comes to mind (e.g., *sweet*). Reusing response words as cues for subsequent individuals has enabled researchers to map out networks of WA[16]. This research has substantiated the idea that information in the human mind is organized into *concepts* – sets of objects, events, or abstract entities[17]. In WA networks, concepts are represented as coherent clusters of words[18].

Being derived from collective natural language use, network models based on WE may tap collective conceptual structures. If so, the clusters that emerge from WE networks should strongly resemble the clusters that emerge from WA networks. Accordingly, we test the correspondence between clusters that emerge from WE and WA networks based on the same constituent words. A strong correspondence between WE and WA clusters would justify WE as a means to study the conceptual organization of the human mind and the concepts contained therein.

In Study 1, we determine the *intrinsic* convergence between WA and WE clusters by comparing the respective networks without parameterization. In Study 2, we determine the *maximal* convergence between WA and WE clusters by imposing various network restrictions. Finally, in Study 3, we determine the *practical* convergence between WA and WE clusters by simulating various experimental conditions. Together, these three studies introduce and validate a novel set of methods for modelling concepts using WE. Accordingly, this article concludes with guidelines and directions for concept modelling across scientific disciplines.

**Study 1: Intrinsic Convergence**

Without imposing restrictions or pre-screening words, three WA networks were derived from self-reported WA in English, Dutch, and Japanese. Concurrently, three congruent WE networks were derived from pre-trained WE vectors in English, Dutch, and Japanese. Potential concepts within each network were identified using cluster analysis (see Methods).

The degree of overlap between WA and WE was determined using two metrics: Informational Convergence and Semantic Convergence. *Informational Convergence* (IC) represents the degree to which WE and WA clusters contain the same words. This was operationalized using the following information-theoretic equation[19], where H(A) represents the Shannon Entropy for a partition of WA, and H(A|E) represents the conditional entropy of a WA partition given a WE partition:

$$IC(WA, WE) = \left(\frac{2 \times [H(A) - H(A|E)]}{H(A) + H(E)}\right) \times 100$$

Informational Convergence assumes a value between 0% and 100%, equaling 100% only when partitions are identical, and 0% when partitions are independent.

*Semantic Convergence* represents the degree to which clusters convey the same meaning. This was calculated by comparing position of words in WE vector space associated with WA clusters and WE clusters. The summation of a set of WE vectors produces a new vector that represents the meaning that is common to the set[20]. Moreover, the pre-trained word vector that is nearest to this artificial vector can be used as a label for that set. Based on these two principles, the meaning of each WA and WE cluster was calculated by summing the associated WE vectors within that cluster. This produced a single category vector and a unique label for each cluster (see Fig. 1a). A pair of WE and WA clusters denoted similar categories if their associated vectors were above an established threshold[21] of cosine similarity (Fig. 1b).

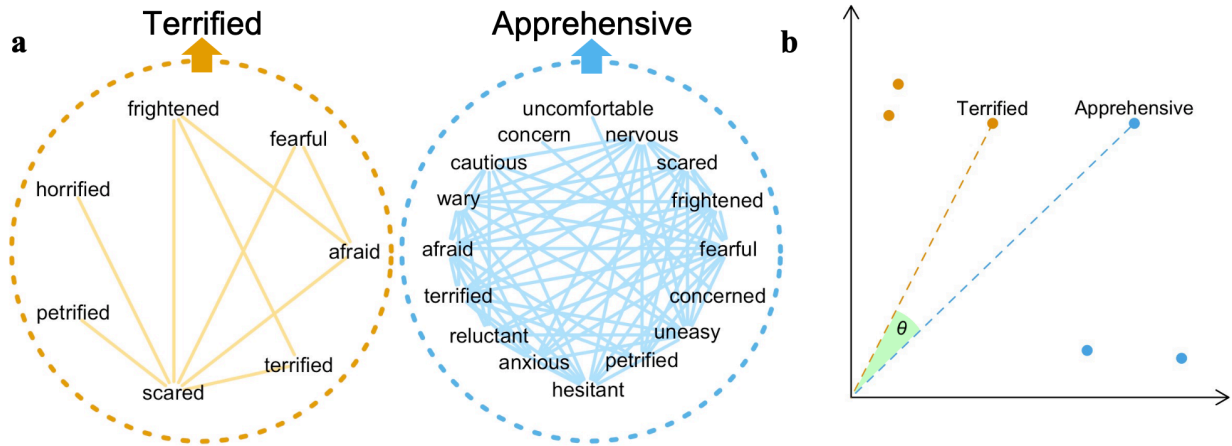

**Fig. 1 | Procedure for comparing the meaning of Word Association (WA) and Word Embedding (WE) clusters.** (**a**) Summation of words within clusters produced a single label and vector for each cluster. An example of a WA (left) and WE (right) cluster are depicted. (**b**) The relative distance of each WA cluster was compared to each WE cluster. Here $Cos_\theta$(Terrified, Apprehensive) > .726 suggesting that the WA "Terrified" cluster has a similar meaning to the WE "Apprehensive" cluster.

Semantic Convergence (SC) was operationalized with the following equation, where $\mathbb{P}$ represents the set of WE-WA cluster pairs above a cosine similarity threshold of .726 without repeating clusters, and $\min\{N_{WA}, N_{WE}\}$ represents the possible number of pairs without repeating clusters.

$$SC(WA, WE) = \left(\frac{\mathbb{P}}{\min\{N_{WA}, N_{WE}\}}\right) \times 100$$

Semantic Convergence assumes a value between 0% and 100%, equaling 100% only when each WE cluster has a unique corresponding WA cluster. Accordingly, Semantic Convergence represents the degree of one-to-one correspondence between WE and WA clusters.

Convergence results from Study 1 are reported in Figure 2. Informational Convergence was high for all three languages, ranging from 76% (Japanese) to 79% (English). Likewise, Semantic Convergence was high for all languages, ranging from 81% (Dutch) to 99% (Japanese). The average cosine similarity between WE and WA categories was well above the similarity threshold, for all languages: English ($M_{cos(\theta)} = .91$, $SD_{cos(\theta)} = .05$), Dutch ($M_{cos(\theta)} = .91$, $SD_{cos(\theta)} = .04$), and Japanese ($M_{cos(\theta)} = .90$, $SD_{cos(\theta)} = .04$). Taken together, these findings provide strong evidence that WE clusters recover the concepts contained in WA clusters.

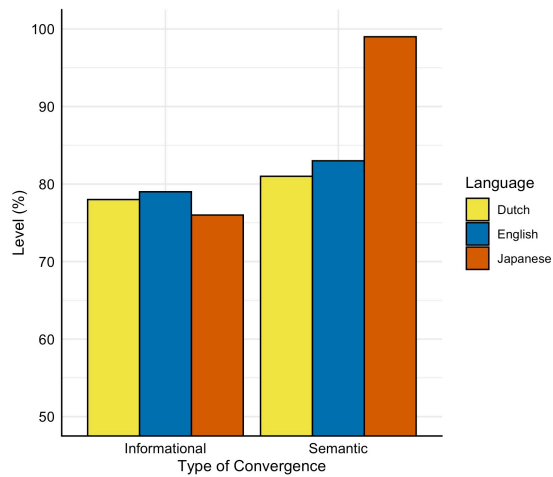

**Fig. 2 | Informational and Semantic Convergence between Word Embedding and Word Association clusters for three languages (Study 1).**

**Study 2: Maximal Convergence**

In an attempt to maximize the convergence between WE and WA clusters, a series of restrictions were applied to make WE and WA more comparable. For these analyses, an alternative sample of English WA was used (see Methods). As in Study 1, a WE matrix of congruent words was created using pre-trained English WE.

First, the effect of vector screening on WE and WA convergence was considered. Specifically, words that appeared to be improperly encoded in the WE vector space were excluded, as demonstrated by isolation from known synonyms (see Methods). As in Study 1, clusters were derived for the resulting networks, and cluster convergence was calculated. Informational Convergence was nearly identical for the screened (76%) and unscreened (77%) data, and similar in scale to Informational Convergence in Study 1. Likewise, Semantic Convergence was nearly identical for screened (74%) and unscreened (75%) data. Thus, screening appeared to have no meaningful effect on WE and WA convergence.

The key difference between WA and WE data is matrix sparsity – WA matrices are inherently sparse, and WE matrices are inherently complete (i.e., cosine values exist between all word vectors). Thus, in an attempt to maximize the convergence between WE and WA clusters, matrix sparsity was systematically manipulated in the screened data. The sparsity of WA was reduced by splitting the data according to part-of-speech boundaries. By replacing a heterogenous WA network with a noun-specific, verb-specific, and adjective-specific network, matrix sparsity was reduced by nearly 50%. Concurrently, sparsity of WE matrices was increased by removing edges between word vectors based on cosine similarity thresholds (see Methods).

Again, clusters were derived from the resulting networks and cluster convergence was calculated (Fig. 3, Chance). Compared to the whole-network baseline (76%), Informational Convergence increased for verb (+3%) and adjective (+8%) models and decreased for the noun model (–3%). Similarly, compared to the whole-network baseline (74%), Semantic Convergence decreased for noun (–9%) and verb (–3%) models, and increased for the adjective model (+4%). Thus, on average, decreasing matrix sparsity resulted in minor improvements in Informational Convergence, and minor decrements in Semantic Convergence.

Finally, the effect of WA strength on subsequent cluster convergence was considered. In the preceding WA networks, the probability of a response given a cue varied considerably (5-95%). Accordingly, concepts based on weak associations could have impaired WE recovery values. To test this notion, WA matrices for each part-of-speech were divided into three classes based on thresholds of associative strength. The minimal probability of a response given a cue word was 10%, 15% and 20% in the Low, Moderate and High classes, respectively. Clusters were derived from the resulting networks and cluster convergence was calculated for each class (Fig. 3).

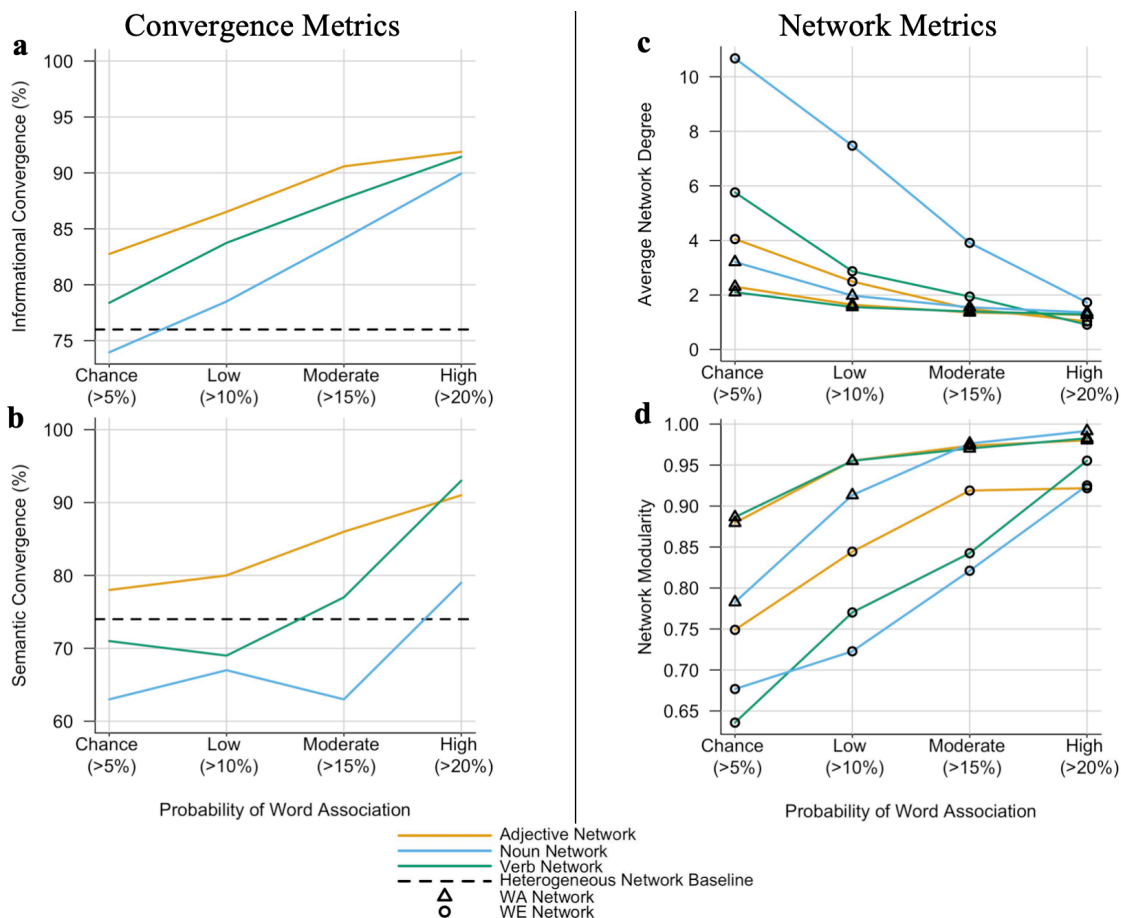

**Fig. 3 | Convergence and network metrics for part-of-speech analyses (Study 2).** (**a**) Informational and (**b**) Semantic Convergence between Word Association (WA) and Word Embedding (WE) clusters as a function of WA probability. (**c**) Average Network Degree and (**d**) modularity as a function of WA probability.

WA strength restriction produced a trend of increasing convergence for all parts of speech. Compared to the unrestricted part-of-speech networks, Informational Convergence increased in High strength networks by 11% (adjectives) to 22% (nouns). Similarly, Semantic Convergence increased by 13% (adjectives) to 22% (verbs). This trend of increasing convergence appeared to correspond with changing network properties (Fig. 2c and 2d). Specifically, as the threshold for associative strength increased, network modularity increased, and average network degree decreased for both WA and WE networks. Thus, increasing associative strength resulted in more precise categories in WE and WA networks, thereby facilitating convergence.

Together, these results suggest that the current state-of-the-art WE can recover mental concepts with around 90% accuracy (at best). Recovery is maximal for the most robust WA concepts, in part due to the increased specificity of these concepts. Recovery is diminished for noun-specific concepts, in part due to the pronounced interconnectedness of noun-specific categories in WE networks.

**Study 3: Practical Convergence**

In practice, concepts are modelled in isolation. For instance, researchers may seek to determine the basic categories of human emotion by focusing solely on emotion words. This introduces two parameters which may affect concept recovery: i) restrictive sampling, ii) potential noise – i.e., concept irrelevant words. Accordingly, the influence of these two factors on Informational Convergence was tested, using the English WA dataset employed in Study 2.

First, 5 to 50 concepts were incrementally sampled from the High strength WA networks for each part-of-speech (1000 samples per block). For each sample, a corresponding WE network was derived, and Informational Convergence was calculated using a sample size optimized WE threshold (see Methods). The resulting curve represents the average Informational Convergence per sample (Fig. 3a). In line with the whole network results, the average Informational Convergence remained above 90% for all parts-of-speech. Thus, restrictive sampling does not appear to diminish the convergence between WE and WA.

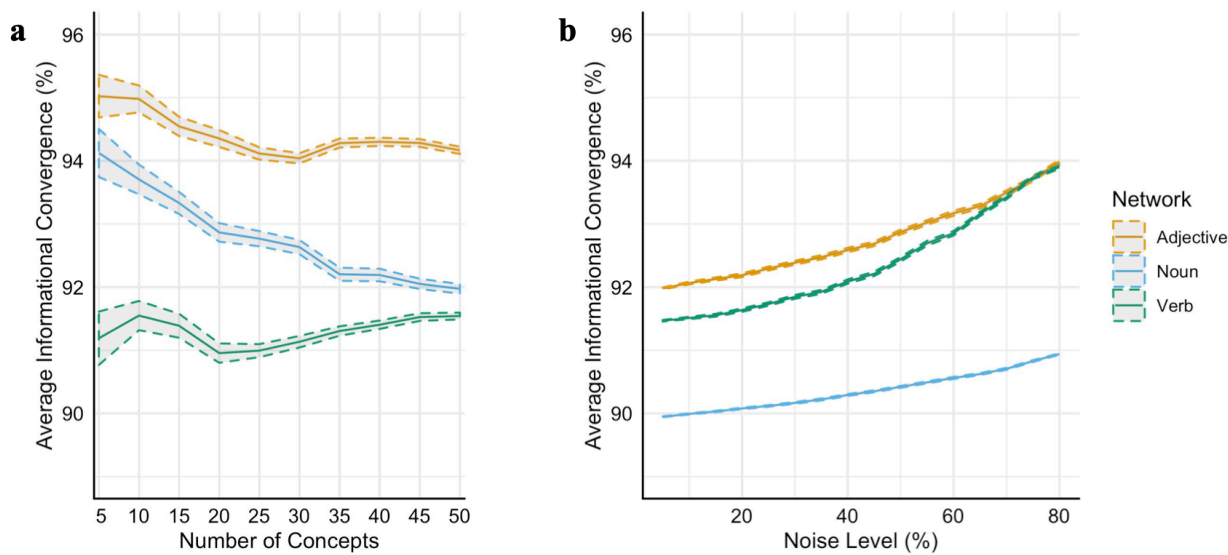

**Fig. 4 | Convergence under practical conditions.** (**a**) Impact of concept sampling on Informational Convergence between WE and WA clusters. (**b**) Impact of noise introduction on Informational Convergence between WE and WA clusters.

Second, noise was incrementally introduced into WE networks (Moderate strength) by randomly incorporating weakly associated words (<1% response-cue probability). Clusters were extracted from the resulting WE networks and Informational Convergence with noise-free WA clusters was calculated (500 samples per block). The resulting curve represents the impact of noise on Informational Convergence (Fig. 3b). Surprisingly, introducing noise was found to increase convergence for all networks – possibly by providing attractors for otherwise spurious connections. Thus, incorporating noise does not appear to diminish convergence between WE and WA.

**Implications and Applications**

Taken together, these results suggest that WE are capable of recovering human concepts with a high degree of accuracy – across languages and sampling conditions. Because WE are based on the natural language use of large groups (in this case, millions) of people, WE concept mapping represents a new methodology situated between self-report and naturalistic observation. Unlike self-report, WE concept mapping is highly scalable and free from idiosyncratic response biases. Unlike observational studies, WE concept mapping is time/cost effective and free from subjective coding biases.

Compared to traditional methods of concept mapping, WE affords four enticing qualities:

i)     Scale: WE models can contain all of the words within a language or set of languages[22]
ii)    Depth: WE preserve group biases while minimizing idiosyncratic biases[23]
iii)   Efficiency: Concepts can be mapped using pre-trained WE almost instantly[1]
iv)    Regularity: WE vectors preserve Euclidian space, enabling interpretable transformations[20]

These qualities combine to make WE concept mapping a valuable tool for any concept-based projects – from identifying weaknesses in philosophical concepts to determining basic categories of perception.

As demonstrated herein, concepts can be recovered from pre-trained WE vectors by applying cluster analysis. Specifically, agglomerative clustering can be used when mapping large-scale networks of concepts (> 1000 anticipated concepts). When mapping fewer concepts, random-walk based clustering is recommended, after inducing matrix sparsity using cosine similarity thresholds (see Methods).

After modelling a set of concepts in WE space, a number of techniques can be applied to WE vectors to uncover, define, assess and relate concepts. Basic arithmetic operations can combine or refine the meaning of WE vectors[20]. For instance, vector addition reveals the common meaning of words (rose + daisy ≈ flower), while vector subtraction reveals the unique meaning of words (rose – daisy ≈ surged). Projection Matrices can uncover hierarchical relationships among WE vectors[24]. Semantic Projection[25] can selectively impose dimensions of meaning in WE space. Moreover, Diachronic Word Embedding[26] can track semantic changes over time. Combined with cluster analysis – which can identify coherent concepts – these techniques can be used to answer a wide variety of concept-related questions (Table 1).

It is difficult to overstate that WE concept mapping can prove to be useful in *any* domain where concepts are studied or utilized. In the physical sciences, WE may be used to uncover ambiguous or contradictory concepts that need development. In the life sciences, WE can uncover biases in diagnostic categories. In the social sciences, WE can be used to develop and validate scales. Beyond any single subject area, WE can help us find links between field-specific concepts, thereby facilitating communication between disciplines and fostering solutions by analogy.

| Define Concepts | Assess Concepts | Relate Concepts | Uncover Concepts |
|---|---|---|---|
| What are the core features of A? | To what degree is A elaborated? | Is A distinct from B? | What semantic regions lack words? |
| What dimensions define A? | To what degree is A isolated? | Did A evolve out of B? | What term embodies a diverse set? |
| What are extreme forms of A? | How does A change over time? | Do A and B share core features? | What concepts are analogical to A? |
| What terms constitute a dictionary of A? | How is A represented in different languages? | Do A and B share dimensions of meaning? | How does A translate into other parts-of-speech? |

**Table 1 | Types of questions that can be answered using Word Embeddings.** "A" and "B" denote concepts of interest.

**References and Notes:**
1. Mikolov, T., Grave, E., Bojanowski, P., Puhrsch, C., & Joulin, A. Advances in pre-training distributed word representations. Preprint at https://arxiv.org/abs/1712.09405 (2017).
2. Hollis, G., Westbury, C. & Lefsrud, L. Extrapolating human judgments from skip-gram vector representations of word meaning. *Q. J. Exp. Psychol*. **70**, 1603–1619 (2016).
3. Hollis, G. & Westbury, C. The principals of meaning: Extracting semantic dimensions from co-occurrence models of semantics. *Psychon. Bull. Rev*. **23**, 1744–1756 (2016).
4. Heyman, T. & Heyman, G. Can prediction-based distributional semantic models predict typicality?. *Q. J. Exp. Psychol*. **72**, 2084–2109 (2019).
5. Pereira, F., Gershman, S., Ritter, S. & Botvinick, M. A comparative evaluation of off-the-shelf distributed semantic representations for modelling behavioural data. *Cogn. Neuropsychol*. **33**, 175–190 (2016).
6. Baroni, M., Dinu, G. & Kruszewski, G. Don't count, predict! A systematic comparison of context-counting vs. context-predicting semantic vectors. *Proc. Assoc. Comput. Linguist*. **52**, 238–247 (2014).
7. Mandera, P., Keuleers, E. & Brysbaert, M. Explaining human performance in psycholinguistic tasks with models of semantic similarity based on prediction and counting: A review and empirical validation. *J. Mem. Lang*. **92**, 57–78 (2017).
8. Bhatia, S. & Walasek, L. Association and response accuracy in the wild. *Mem. Cogn*. **47**, 292–298 (2019).
9. Bhatia, S. Predicting risk perception: New insights from data science. *Manag. Sci*. **65**, 3800–3823 (2019).
10. Richie, R., Zou, W. & Bhatia, S. Semantic representations extracted from large language corpora predict high-level human judgment in seven diverse behavioral domains. Preprint at https://psyarxiv.com/g9j83 (2019).
11. Bhatia, S. Associative judgment and vector space semantics. *Psychol. Rev*. **124**, 1–20 (2017).
12. Kozlowski, A. C. & Taddy, M. The geometry of culture: Analyzing meaning through word embeddings. Preprint at https://arxiv.org/abs/1803.09288 (2018).
13. Garg, N., Schiebinger, L., Jurafsky, D. & Zou, J. Word embeddings quantify 100 years of gender and ethnic stereotypes. *Proc. Natl. Acad. Sci*. **115**, 3635–3644 (2018).

14. Hellrich, J. Word embeddings: Reliability and semantic change. (Friedrich Schiller University, Jena, 2019).
15. De Deyne, S., Navarro, D. J., Storms, G., Better explanations of lexical and semantic cognition using networks derived from continued rather than single-word associations. *Behav. Res. Methods*. **45**, 480–498 (2013).
16. Steyvers, M. & Tenenbaum, J. The large-scale structure of semantic networks: Statistical analyses and a model of semantic growth. *Cogn. Sci.* **29,** 41–78 (2005).
17. Mareschal, D., Quinn, P. & Lea, S. *The making of human concepts*. New York: Oxford University Press (2010).
18. De Deyne, S. & Storms, G. Word associations: Network and semantic properties. *Behav. Res. Methods*. **40**, 213–231 (2008).
19. Danon, L., Diaz-Guilera, A., Duch. J. & Arenas, A. Comparing community structure identification. *J Stat Mech*. **9**, 1–10 (2005).
20. Mikolov, T., Sutskever, I., Chen, K., Corrado, G. & Dean, J. Distributed representations of words and phrases and their compositionality. Preprint at https://arxiv.org/abs/1310.4546 (2013).
21. Rekabsaz, N., Lupu, M. & Hanbury, A. Exploration of a threshold for similarity based on uncertainty in word embedding. *Lect. Notes Comput. Sci.* **10193**, 396–409 (2017).
22. S. Ruder, S., Vulić, I., & Søgaard, A. A survey of cross-lingual word embedding models. *J Artif. Intell. Res*. **65**, 569–631 (2019).
23. Caliskan, A., Bryson, J. J., & Narayanan, A. Semantics derived automatically from language corpora contain human-like biases. *Science*. **356**, 183–186 (2017).
24. Fu, R. et al. Learning semantic hierarchies via word embeddings. *Proc. Assoc. Comput. Linguist*. **52**, 1199–1209 (2014).
25. Grand, G., Blank, I. A., Pereira, F. & Fedorenko, E. Semantic projection: Recovering human knowledge of multiple, distinct object features from word embeddings. Preprint at https://arxiv.org/abs/1802.01241 (2018).
26. Kutuzov, A., Øvrelid, L., Szymanski, T. & Velldal, E. Diachronic word embeddings and semantic shifts: A survey. Preprint at https://arxiv.org/abs/1806.03537 (2018).

## Methods
**Datasets**

For the Word Embedding (WE) networks, I utilized publicly available pre-trained word vectors published by Mikolov et al.[1, 27]. Each model contained 2 million 300-dimensional word vectors derived using the FastText algorithm, a variation of continuous bag-of-words (CBOW). Sub-word information was incorporated on the basis of n-grams (length = 5), with a window size of 5 and 10 negatives, and a step size of .05. The English model was trained on a Common Crawl corpus comprised of English text from 2.96 billion webpages. Given the relative paucity of text in other languages, the Dutch and Japanese models were trained on corpora which combined text from both Common Crawl and Wikipedia, in their respective languages.

For the Word Association (WA) networks, I utilized four datasets of WA norms – three publicly available, and one available upon request. In each WA dataset, participants were provided with a cue word and asked to respond with the first word that came to mind. The SWOW English WA dataset consists of 12,218 American English cue words and 3,665,100 responses that were provided by 83,864 participants[28]. The SWOW Dutch WA dataset consists of 1,424 Dutch cue words and 381,909 responses that were provided by 10,292 participants[29]. In both SWOW datasets, each participant provided three responses for a set of cue words (up to 30), with each cue word accumulating responses from a minimum of 115 (English) or 300 (Dutch) participants. The Japanese WA Database (JWAD) consists of 2,099 basic Japanese kanji cue words and 53,275 responses provided by 1,486 native Japanese speakers[30]. On average, 50 participants ($SD = 1$) provided responses for a subset of 100 cue words. These datasets formed the basis of Study 1.

Study 2 and 3 were based on the USF English WA dataset. This dataset consists of 5,019 English cue words and nearly 750,000 responses provided by more than 6,000 participants (precise values not reported)[31]. The USF dataset was selected for fine grained analysis due to several enticing features. Firstly, the USF dataset alone contained normed cues – i.e., popular responses provided for cues early in the study were administered as cues to subsequent groups of participants in order to keep the semantic network more focused. Secondly, the USF dataset alone contained manually coded parts-of-speech. Finally, this dataset provided the most reliable WA probability values, as these values were based on both a large number of participants per cue ($M = 149$, $SD = 15$) and a large number of cues per participant (100-120 words).

**Dataset Preparation**
**1. Word Association Edge Weight Calculation**
The forward strength of association (FSA) between words represents the probability of producing a response (R) given a cue (C). When a response word is used as a cue ($C_R$), the backwards strength of association (BSA) represents the probability of producing the original cue (C) as a response to $C_R$. Both FSA and BSA were calculated by considering the proportion of participants who provided a particular response following a particular cue. For example, of the 283 times that "outer" was presented as a cue in the SWOW English dataset, it was followed by the response "space" 49 times – thus, FSA = .17. Conversely, of the 296 times that "space" was presented as a cue in this dataset, it was followed by the response "outer" 22 times – thus, BSA = .07. Network edge weights for the WA datasets were calculated by averaging BSA and FSA.

**2. Word Association Pre-Screening**
To be included in the WA networks, word-pairs needed to satisfy four conditions. Firstly, each word must be present in the WE corpus of the respective language. Secondly, the word pair must have both an FSA and BSA value, in order to calculate balanced network edge weights. Thirdly, the associative probability of the word pair must be above 5%. This condition was applied to exclude spurious or idiosyncratic associations. Finally, each word must be identified as either a noun, verb, adjective, or adverb. Functional parts-of-speech (e.g. pronouns) were excluded as these are both context-insensitive and difficult to compare across languages. With the exception of USF (manually tagged), part-of-speech tagging was carried out using language-specific taggers in Python – Nagisa (Japanese), spaCy (Dutch), and NLTK (English). The properties of the screened WA datasets are reported in Table 2.

|  | Word Pairs | Unique Words | Nouns | Verbs | Adjectives | Adverbs |
| --- | --- | --- | --- | --- | --- | --- |
| SWOW English | 17,952 | 7,006 | 53% | 30% | 14% | 3% |
| SWOW Dutch | 19,362 | 7,802 | 64% | 18% | 15% | 3% |
| JWAD Japanese | 1,137 | 717 | 72% | 17% | 9% | 2% |
| USF English | 9,961 | 4,899 | 69% | 17% | 13% | 1% |

**Table 2 | Properties of screened Word Association datasets.**

**3. Word Embedding Screening (Study 2)**
WE vectors that were distant from known synonyms were identified as improperly encoded and excluded from analyses in Study 2 and 3. Synonyms were aggregated for each vector by searching Myriam Webster's Thesaurus online using a Python API. The cosine distance between each WE vector and its associated synonyms was calculated. Vectors with a cosine similarity of .692 or less for all associated synonyms were deemed improperly encoded. This similarity threshold was established in previous research[21]. This screening method resulted in the exclusion of 130 nouns, 22 verbs, 5 adjectives, and 2 adverbs.

**Deriving Clusters**

For all three studies, clusters were derived from WA networks using Infomap, a random walk algorithm based on the Map equation[32]. The cluster analysis procedure varied between studies for WE networks due to changing matrix properties. In Study 1, and the heterogeneous network analysis in Study 2, adjacency matrices were saturated with edges between all nodes. Accordingly, Infomap could not differentiate the community structure of the network. Thus, for these analysis, Infomap was replaced with an agglomerative clustering algorithm[33]. The optimal level of resolution was determined using the Dynamic Tree Cut algorithm[34], with a minimum cluster size of 2, a maximum joining height of .99 and a minimum split height of 0. Once sparse matrices had been induced for WE networks using thresholding (Study 2 onwards), Infomap was used to derive clusters for these networks.

**Thresholding WE Networks**

To induce sparse matrices in Study 2 and 3, network edges that were below specific cosine thresholds in WE were set to 0. In Study 2, these thresholds were determined by calculating the average cosine similarity for all WE word pairs within specific WA probability ranges: <1%, 5-10%, 10-15%, 15-20%, and >20%. Separate thresholds were calculated for each part-of-speech network. In order of increasing probability ranges, the resulting cosine thresholds were similar for nouns (.65, .71, .72, .74, .77), verbs (.67, .72, .74, .75, .78), and adjectives (.68, .75, .77, .79, .81). In the part-of-speech networks without imposed strength classes, only values below the 5% probability threshold were set to zero. In networks with imposed strength class, the applied threshold corresponded to the WA probability limit. For instance, cosine values less than .77 were set to 0 in the WE noun network when testing convergence with High class WA (> 20% chance of response given cue).

In Study 3, optimal thresholds were calculated for networks of different sizes (5 to 50 concepts) using bootstrapped sampling. 1000 concept sets were derived by randomly sampling WA word pairs at each size interval, in increments of 5 concepts. An increasing cosine threshold (.65 to .85) was applied to the corresponding WE matrices and the resulting informational convergence was recorded. The cosine threshold that maximized Informational Convergence was selected for each network. The optimal cosine threshold increased as a function of the number of concepts, for adjectives (.73 to .78), nouns (.68 to .74), and verbs (.71 to .77). These values may be useful for researchers seeking to model a known number of concepts using WE.

**Network Parameters**

The modularity of a network represents the degree of separation between different partitions of that network. The average degree of a network represents the average number of edges between vertices. Modularity and degree were calculated using the igraph[35] modularity and degree functions.

**Competing interests**

The author declares no competing interests.

**Code Availability**

All code used herein will be made available online upon publication.

**Data Availability**

The pre-trained Word Embeddings used herein are available at https://fasttext.cc. Word association norms used in this study are available at http://w3.usf.edu/FreeAssociation/ and http://www.smallworldofwords.com. The JWAD dataset is available upon request from the original author[28].

## Methods References

27. Grave, E., Bojanowski, P., Gupta, P., Joulin, A. & Mikolov, T. Learning word vectors for 157 languages. Preprint at https://arxiv.org/abs/1802.06893 (2018).
28. De Deyne, S., Navarro, D., Perfors, A., Brysbaert, M. & Storms, G. The 'small world of words' English word association norms for over 12,000 cue words. *Behav. Res. Methods.* **51,** 987–1006 (2019).
29. De Deyne, S. & Storms, G. Word associations: Norms for 1,424 Dutch words in a continuous task. *Behav. Res. Methods.* **40,** 198–205 (2008).
30. Joyce, T. Constructing a large-scale database of Japanese word associations. *Glottometrics*. **10**, 82–98 (2005).
31. Nelson, D., McEvoy, C. & Schreiber, T. The University of South Florida free association, rhyme, and word fragment norms. *Behav. Res. Meth. Instrum. Comput.* **36,** 402–407 (2004).
32. Rosvall, M. & Bergstrom, C. T. Maps of information flow reveal community structure in complex networks. *Proc. Natl. Acad. Sci.* **105**, 1118–1123 (2008).
33. Müllner, D. Fastcluster: Fast hierarchical, agglomerative clustering routines for R and Python. *J. Stat. Softw*. **53**, 1–18 (2013).
34. Langfelder, P., Zhang, B. & Horvath, S. Defining clusters from a hierarchical cluster tree: The Dynamic Tree Cut package for R. *Bioinformatics*. **24**, 719–720 (2008).
35. Csardi, G. & Nepusz, T. The igraph software package for complex network research. *Inter. J. Complex Syst.* **1695**, 1–9 (2006).